\documentclass{article}

\usepackage{microtype}
\usepackage{graphicx}
\usepackage{booktabs} 

\usepackage{hyperref}

\usepackage[accepted]{icml2023}

\usepackage{amsmath}
\usepackage{amssymb}
\usepackage{mathtools}
\usepackage{amsthm}

\allowdisplaybreaks 

\usepackage{amsmath,amssymb,amsfonts} 
\usepackage{upgreek}
\usepackage{nicefrac}

\DeclareMathOperator*{\argmin}{arg\,min}

\usepackage{setspace}
\usepackage{marvosym} 
\usepackage{xspace}

\usepackage{subfigure}

\usepackage{blindtext}

\usepackage{tikz-qtree}
\usepackage{tikz-cd}

\usepackage{csquotes}
\usepackage{enumitem}

\usepackage{algorithmic}
\usepackage[algo2e, ruled, lined, linesnumbered, titlenumbered]{algorithm2e}
\newcommand{\InOut}{\SetKwInOut{Input}{Input}\SetKwInOut{Output}{Output}}
\SetKwComment{Comment}{$\triangleright$\ }{}

\usepackage[acronym]{glossaries}

\newacronym{mcs}{MCS}{Monte Carlo Simulation}

\newcommand*{\constraintref}[1]{\hyperref[#1]{Constraint~\ref*{#1}}}
\newcommand*{\inequalityref}[1]{\hyperref[#1]{Inequality~\ref*{#1}}}
\newcommand*{\opref}[1]{\hyperref[#1]{Optimization Problem~\ref*{#1}}}

\theoremstyle{plain}
\newtheorem{theorem}{Theorem}[section]

\newtheorem{corollary}[theorem]{Corollary}
\theoremstyle{definition}
\newtheorem{definition}[theorem]{Definition}

\theoremstyle{remark}

\usepackage[disable,textsize=tiny]{todonotes}

\icmltitlerunning{Selecting Models based on the Risk of Damage Caused by Adversarial Attacks}

\begin{document}

\newcommand{\R}{\mathbb{R}}
\newcommand{\N}{\mathbb{N}}
\newcommand{\E}{\mathbb{E}}
\renewcommand{\P}{\mathbb{P}}
\newcommand{\M}{\mathcal{M}}
\newcommand{\A}{\mathcal{A}}
\newcommand{\Rn}{\R^n}
\newcommand{\eqdot}{.}
\newcommand{\X}{\mathcal{X}}
\newcommand{\CSR}{\text{CSR}}
\newcommand{\CSRt}{\CSR^{\text{true}}_{\A, m}}
\newcommand{\CSRem}[1]{\CSR^{\text{emp}}_{\A, #1, \X}}
\newcommand{\CSRe}{\CSRem{m}}
\newcommand{\Kda}{\mathcal{K}^d_\A}
\newcommand{\cardinality}[1]{\left\lvert#1\right\rvert}
\newcommand{\dam}{\text{damage}}
\newcommand{\dete}{\text{detection}}
\newcommand{\adv}[1][M]{\Uppi^{#1}_{\A}}
\newcommand{\advnoA}{\Uppi^{M}}
\newcommand{\pdam}{P^\text{dam}}
\newcommand{\pdamhat}{\widehat{{P}}^\text{dam}}
\newcommand{\cdam}{C^{\text{dam}}}
\newcommand{\Risk}{R}
\newcommand{\Riske}{R^\text{emp}}
\newcommand{\Riskam}{\Risk_{\A, m}}
\newcommand{\Ex}{\text{Ex}}
\newcommand{\Dma}[1][M]{\Psi_{\A, #1}}
\newcommand{\Dmahat}[1][M]{\hat{\Psi}_{\A, #1}}
\renewcommand{\H}{H}
\newcommand\Tau{\mathrm{T}}
\newcommand{\dsa}{d_{\A}}
\newcommand{\F}{\mathcal{F}}
\newcommand{\dct}{dct}

\newcommand{\limitA}{Small Sample Size\xspace}
\newcommand{\limitB}{Model-Specific Estimates\xspace}

\newcommand{\modelCarmon}{\texttt{Carmon-Semi}\xspace}
\newcommand{\modelRice}{\texttt{Rice-Overfit}\xspace}
\newcommand{\modelEngstrom}{\texttt{Engstrom-Robust}\xspace}
\newcommand{\modelBaseline}{\texttt{Baseline}\xspace}

\newtheorem{intern}{Internal Remark.}
\twocolumn[
\icmltitle{Selecting Models based on the Risk of Damage Caused by Adversarial Attacks}


\icmlsetsymbol{equal}{*}

\begin{icmlauthorlist}
\icmlauthor{Jona Klemenc}{equal,neurocat}
\icmlauthor{Holger Trittenbach}{equal,neurocat}
\end{icmlauthorlist}

\icmlaffiliation{neurocat}{neurocat, Berlin, Germany}

\icmlcorrespondingauthor{Jona Klemenc}{jona.klemenc@neurocat.ai}
\icmlcorrespondingauthor{Holger Trittenbach}{holger.trittenbach@neurocat.ai}

\icmlkeywords{Adversarial Attack, Robustness, Risk Score}

\vskip 0.3in
]



\printAffiliationsAndNotice{\icmlEqualContribution} 

\begin{abstract}

Regulation, legal liabilities, and societal concerns challenge the adoption of AI in safety and security-critical applications.
One of the key concerns is that adversaries can cause harm by manipulating model predictions without being detected.
Regulation hence demands an assessment of the risk of damage caused by adversaries.
Yet, there is no method to translate this high-level demand into actionable metrics that quantify the risk of damage.

In this article, we propose a method to model and statistically estimate the probability of damage arising from adversarial attacks.
We show that our proposed estimator is statistically consistent and unbiased.
In experiments, we demonstrate that the estimation results of our method have a clear and actionable interpretation and outperform conventional metrics.
We then show how operators can use the estimation results to reliably select the model with the lowest risk.

\end{abstract}

\section{Introduction}\label{sec:intro}

Adversarial perturbations are a security risk since an adversary can use them to alter machine learning model predictions without being noticed by a human~\cite{Yuan2019-qa, Ren2020-bc}.
For instance, think of an upload filter that uses a machine learning model to identify prohibited material on social media platforms, such as copyright-protected images or hate speech.
By perturbing prohibited content, an adversary may bypass an upload filter even though the content appears identical to the original prohibited content to the human observer.
When machine learning models are deployed in systems that take or enable action in physical environments, the security risks can result in safety hazards~\cite{Deng2020-ld}.
An example is autonomous driving, where a false classification, e.g., of a stop sign, can severely damage property and life.
Given these potentially severe consequences, regulators have made clear that one must treat risk from adversaries seriously.
Broad regulation, such as the \enquote{EU AI Act}~\cite{eu-ai-act}, as well as domain-specific norms, such as safety standards for autonomous vehicles (e.g., ISO 21448 (SOTIF)~\cite{British_Standards_Institution2022-os}, UL 4600~\cite{Underwriters_Laboratories_Inc2022-ry}, and ISO PAS 8800~\cite{International_Organization_for_Standardization_undated-le}), explicitly require an assessment of the risk arising from adversarial attacks.
Consequently, operators of machine learning models strive to manage the risk that stems from adversarial perturbations~\cite{Piorkowski2022-zy}. 
While the motivation and intentions are clear, neither the regulation nor the academic literature currently provides sufficient technical guidelines on how to assess the \enquote{risk of adversarial attacks}.

A risk assessment of adversarial attacks requires a reliable estimate that an adversary causes damage.
Intuitively, this \textit{probability of damage} describes how likely an adversary can find perturbations that go undetected and alter model predictions.
The probability of damage hence depends on the capabilities of an adversary and the effectiveness of measures put in place to detect adversarial perturbations.
Previous work has focused on evaluating adversaries by comparing adversarial attacks~\cite{Yuan2019-qa, Ren2020-bc} with each other, e.g., by a per-attack drop in accuracy~\cite{Brendel2020-is}.
These evaluations do not consider the likelihood that measures detect adversarial perturbations.
Instead, evaluations assume a threshold~\cite{croce2021robustbench, Maho2021-qs} on the perturbation size beyond which perturbations are detected with certainty.
This is a stark simplification.
Practical experience shows that a suitable threshold is hard to find or may not exist.
Other proposed evaluation metrics that compare perturbation sizes~\cite{Carlini2019-yj} are not viable alternatives as they are either prone to outliers or suffer from statistical bias.
It is an open question of how to reliably estimate the probability of damage and how to use the estimate to select machine learning models.

In this article, we propose a statistical approach to assessing the risk caused by adversarial attacks.
Our main contribution is a model-agnostic estimator for the probability of damage that is statistically unbiased and consistent.
Our estimator explicitly considers the probability of detecting perturbations instead of assuming a hard threshold.
As it turns out, calculating estimates for the probability of detection efficiently is challenging for large sample sizes (see \autoref{sec:adversarial-risk-score}).
Hence, we propose a strategy to make estimates efficient (see \autoref{subsec:inferred_damage_prob}), even when querying the detector, e.g., a human in the loop, is out of reach (see \autoref{sec:relative-estimation}).
The estimates resulting from our method allow comparing different models with each other to select the model with the lowest risk of damage caused by adversarial attacks.
Our experiments demonstrate that our estimator is more reliable than existing metrics in adversarial robustness benchmarks.

\section{Notation}

Let $\mathcal{X}$ be a data space, and $\mathbf{X} = \langle x_1,\dots, x_I \rangle \subseteq \mathcal{X}$ a sample of $I$ observations.
A machine learning model $M \colon \mathcal{X} \rightarrow \mathcal{O}$ is a function that maps the data space to a prediction space $\mathcal{O}$.
$\mathcal{O}$ differs depending on the model task, e.g., for image classification it might be the space of logit scores or class labels; for object detection it might be the space of bounding boxes and classification scores.
When there are multiple models, we index them as $M_1, \dots, M_J$.

We further define a \textit{ground truth} as a function $g \colon \mathcal{X} \rightarrow \mathcal{O}$ that assigns each observation a value of the prediction space that is considered \enquote{correct} by some gold standard.
We say a prediction is \enquote{incorrect} if $M(x) \neq g(x)$.
We say that two model predictions \enquote{disagree} for a pair of observations if $M(x) \neq M(x'), x, x' \in \mathbf{X}$.
Herein, we do not specify further what it means to disagree since the specifics usually depend on the model task and application.
For instance, in one case, one would say two models disagree if the argmax of their logit outputs are different; in another case, they disagree if the ranking of their top-k logit scores differs.

We use $P_{S}$ to denote probability distribution functions over some space $S$ with the corresponding probability density function $dP_{S}$.
The hat notation indicates empirical estimates, e.g., an empirical distribution $\hat{P}$.

We say an estimator of a probability function is \emph{unbiased} if the mean of the sampling distribution of the estimator is equal to the true probability.
An estimator that converges to the estimated value with increasing sample size is \emph{consistent}~\citep[see][p.~351]{cramer1999mathematical}.
\section{Fundamentals and Related Work}
\label{sec:related-work}

We expect operators to assess the risk of using a machine learning model in a security-critical application as an expected value of the total damage, i.e., the product of the occurrence probability of damage when operating a model $\pdam(M)$ and the expected size of the damage $\cdam$.
\begin{equation}
    \label{eq:operational-risk}
    \text{Operational Risk}(M) = \pdam(M) \times \cdam
\end{equation}
We assume that $\cdam$ is constant and independent of whether the operator relies on machine learning or any other system, e.g., a human in the loop.
We focus on the risk of a malicious adversary seeking to intentionally manipulate predictions in a way that can inflict damage.
We do not consider other categories of machine-learning-related security and privacy risks, such as the risks of model stealing~\cite{tramer2016stealing} and model inversion~\cite{fredrikson2015model}.

\textbf{Adversarial Risk.} The adversarial machine learning community often uses the term \enquote{risk} in the sense of \emph{Adversarial Risk (AR)}.
AR is concerned with the empirical risk estimation of a model under small perturbations.
Formally, the adversarial risk is the expected loss over perturbations within a neighborhood $\mathcal{N}_\epsilon(x)$
\begin{equation}
\label{eq:adversarial-risk}
    \text{AR}(M) = \mathbb{E}_{x \sim \mathcal{X}}\left[\sup_{x'\in\mathcal{N}(x)}l(M(x'), g(x))\right]
\end{equation}
with loss function $l$~\cite{Uesato2018-dx}.
With $\epsilon\!=\!0$, \autoref{eq:adversarial-risk} reduces to the standard empirical risk.
Typically, the neighborhood $\mathcal{N}$ is constrained to an epsilon ball around $x$, i.e., $\mathcal{N}(x)\!\equiv\! \mathcal{B}_{\epsilon}(x)$ for a fixed perturbation budget $\epsilon$.
Details of this definition differ across the literature~\cite{Pydi2022-qi}.
For instance, instead of the loss against a ground truth, one may compute the loss of prediction change $l(M(x), M(x'))$ if the ground truth is unknown~\cite{Diochnos2018-zh}.

Related research focuses on average risk under random~\cite{Levy2021-vy,Rice2021-vn} and natural perturbations~\cite{Pedraza2021-iq,Hendrycks2019-qk,Schwerdtner2020-ny}.
Here, one measure of interest is the \emph{Error-Region Risk (ERR)}~\cite{Diochnos2018-zh}, i.e., the probability that a successful perturbation exists
\begin{equation}
    \text{ERR}(M) = P_{x \in \mathcal{X}}\left[ \exists x' \in \mathcal{B}_\epsilon(x): M(x') \neq g(x') \right]
\end{equation}
However, estimations of ERR rely on random perturbations drawn from a uniform distribution~\cite{Diochnos2018-zh,Webb2018-us}.
Thus, it does not account for \enquote{an explicit and effective adversary}~\cite{Webb2018-us}.

\textbf{Benchmarks.} Adversarial Risk has inspired several benchmarks that compare the effectiveness of individual adversarial attacks and defenses.
The comparisons are based on either the success rate of attacks for a defined perturbation budget $\epsilon$, see \texttt{RobustBench}~\cite{croce2021robustbench} and \texttt{RoBIC}~\cite{Maho2021-qs}, or the average perturbation size attacks require to find successful perturbations, see \texttt{RobustVision}~\cite{Brendel2020-is}.
Such comparisons are \emph{attack-centric}, i.e., effective in evaluating attacks against each other.
However, estimating $\pdam$ requires \emph{model-centric} evaluations that measure how likely an adversary can find successful adversarial perturbations given a set of attacks and perturbation budgets~\citep[cf.][]{Carlini2019-yj}.

Instead of evaluating adversarial risk for a defined budget, one can also plot accuracy or attack success rates against a perturbation budget~\cite{Dong2020-tr,Carlini2019-yj}.
Such plots are useful to investigate the effectiveness of attacks for varying perturbation budgets.
However, visual inspection of plots does not scale beyond the comparison of a few attacks.
It is an open question how one can use these curves to estimate the operational risk; we will come back to this question in~\autoref{sec:experiments-results}.

In summary, there are a variety of evaluation methods that seek to capture some element of adversarial risk.
They provide a set of measures for attack benchmarks but are not immediately applicable to estimate operational risk.
The reason is that they (i) do not account for the probability of detecting adversarial attacks (see \autoref{sec:adversarial-risk-score}) and (ii) are either biased or inconsistent and hence not useful as statistical estimates (see~\autoref{sec:experiments-results}).

\section{Estimating the Probability of Damage}
\label{sec:adversarial-risk-score}

The operational risk of a machine learning model depends on the probability of damage $\pdam$, see \autoref{eq:operational-risk}.
Intuitively, $\pdam$ is the joint probability of finding a successful perturbation (\textit{Succ}) from the space of adversarial perturbations $Adv^M(\mathcal{X})$ and of a detector, e.g., a human in the loop, not detecting it ($\neg$~\textit{Det}).
Formally, we can express $\pdam$ as the joint probability
\begin{align}
    \begin{split}
    \label{eq:joint-success-detection}
    \pdam \coloneqq & P_{x \sim Adv^M(\mathcal{X})}(Succ(x),\ \neg Det(x))  \\
          = & \underbrace{P_{x \sim Adv^M(\mathcal{X})}(Succ(x))}_{ \text{Probability of Attack Success}} \\ 
          & \times \underbrace{P_{x \sim Adv^M(\mathcal{X})}(\neg Det(x) \mid Succ(x))}_{\text{Probability of Detection}}
    \end{split}
\end{align}
Considering this equation, a natural solution to estimating $\pdam$ seems to be \acrlong*{mcs}:
One can draw a sample from the space of adversarial perturbations $Adv^M(\mathcal{X})$ to estimate the probability of attack success $P_{x \sim Adv^M(\mathcal{X})}(Succ(x))$ and then pass the successful perturbations on to a detector to estimate the probability of detection $P_{x \sim Adv^M(\mathcal{X})}(\neg Det(x) \mid Succ(x))$.
However, one must be careful with designing this sampling process -- a naive sampling of random perturbations is not sufficient here.
An adversary is efficient and uses all means available to find successful perturbations.
The adversary does not search through the infinite space of random perturbations but instead uses adversarial attacks, i.e., efficient optimization methods that make heuristic assumptions on the perturbation space.
Thus, naive random sampling underestimates an efficient adversary.
Valid \acrlong*{mcs} requires to \textit{simulate} an adversary, i.e., to run attacks that are in line with the resources and capabilities of the adversary.

While \acrlong*{mcs} is conceptually sound, it has two significant limitations in our context:

\emph{\limitA:} If detection requires a human in the loop, collecting data on detection will inevitably be time-consuming: a detector must inspect each successful perturbation individually.
This may significantly limit the feasible sample size and reduce the estimation quality.

\emph{\limitB:} Probability estimates are specific to the machine learning model and attacks used to search for perturbations.
This is because the conditional probability of detection $P_{x \sim Adv^M(\mathcal{X})}(\neg Det(x) \mid Succ(x))$ relies on $Adv^M(\mathcal{X})$ which is specific to the model and attacks.
Consequently, one must repeat the entire estimation of $\pdam$ for any change in the choice of models or attacks.    Because detection is costly, the estimation becomes impractical for operators who have frequent model iterations and face a research field that frequently produces novel attacks.

Both limitations stand in the way of obtaining an estimate of $\pdam$ in many practical settings.

In this section, we propose a sampling method that overcomes both limitations.
We first establish a rigorous formal framework for the probability of attack success (\autoref{subsec:csr}).
We then turn to the estimation of the probability of detection (\autoref{subsec:inferred_damage_prob}) and introduce a method that is not limited by \emph{\limitA} and \emph{\limitB}.
Lastly, we look at the special case of providing an estimation for $\pdam$ when there is no explicit detector -- a setting that often occurs in academic benchmarks~(\autoref{sec:relative-estimation}).
The result is a formal expression of the probability of damage that has both a clear interpretation and an unbiased and consistent estimator.

\subsection{Estimating the Probability of Attack Success}
\label{subsec:csr}

So far, we have defined $\pdam$ as an estimate over the space of adversarial perturbations $Adv^M(\mathcal{X})$.
However, one typically only has access to a sample $\mathbf{X} \sim \mathcal{X}$, used for training or testing of a machine learning model.
The space of adversarial perturbations is defined implicitly by a pushforward of $\mathcal{X}$ along a function $\advnoA: x \mapsto x'$
\begin{equation}
\label{eq:pdam_by_test_distribution}
\begin{split}
    & P_{x \sim Adv^M(\mathcal{X})}(Succ(x),\ \neg Det(x)) \\ 
    & = P_{x \sim \mathcal{X}}(Succ(\advnoA(x)),\ \neg Det(\advnoA(x)))
\end{split}
\end{equation}
We call $\advnoA$ the \emph{attack strategy}.

One can use a sample $\mathbf{X}$ and an attack strategy $\advnoA$ to \textit{simulate an adversary}.
To illustrate, think of an adversary that wants to evade an upload filter for copyright-protected material.
The adversary would select a copyright-protected image and manipulate it via an attack strategy $\advnoA$ such that the classifier of the upload filter predicts it to be non-protected content.
To estimate the probability that the adversary finds such a successful perturbation, one can simulate the adversary by sampling images from the data distribution of interest to the adversary, here the copyright-protected images, and then manipulate the images with the attack strategy the adversary is expected to use.\\

An attack strategy depends on the choice of search algorithms that an efficient adversary has at their disposal to search for successful adversarial perturbations.
The most effective search algorithms known today are adversarial attacks, i.e., heuristics to find small successful perturbations.
Many adversarial attacks require access to the model, e.g., to obtain inference results or gradients with respect to the model input.
An adversary with restricted model access can only use a subset of adversarial attacks that do not rely on gradient calculations.
We define the \emph{set of applicable adversarial attacks} as
\begin{equation*}
    \A = \{a_k\colon (x, M) \mapsto x'\}_{k \in K}
\end{equation*}
where each $a_k \in \{a_1, a_2, \dots, a_K\}$ is an attack that has access to an input $x$ and potentially restricted access to a model $M$.
In practice, an adversary also may have a limited computational budget, which forces them to select a computationally feasible subset $\A' \subset \A$; the selection of a good subset of adversarial attacks under budget restrictions, however, is a question orthogonal to our current article.\\

An adversary executes an attack strategy as follows.
First, the adversary uses $\mathcal{A}$ to generate candidate perturbations
\begin{equation*}
    Cand_{\A}(x, M) = \{a(x, M) \mid a \in \A \}
\end{equation*}
Out of the candidates, the adversary filters the ones $x' \in Cand_{\A}(x, m)$ where $Succ(x') = \text{True} \land \neg Det(x') = \text{True}$.
Filtering for $Succ(x')$ is straightforward if the adversary can access the model predictions.
For instance, if the adversary is interested in an untargeted misclassification, e.g., changing the prediction from \enquote{copyright-protected} images to any other class, the success filter is
\begin{equation*}
    Cand_{\A}^{Succ}(x, M) = \{x'\! \in \! Cand_{\A}(x, M)\! \mid \! M(x')\! \neq \! M(x)\}
\end{equation*}
where the sample $\mathbf{X}$ are images that initially are classified as $M(x) = $ \enquote{copyright-protected}.

However, filtering is not possible for an adversary because it requires access to $\neg Det(x')$, e.g., a human in the loop.
If an adversary would have access to the detector, then there is no need to use adversarial attacks that minimize perturbation sizes.
The adversary could instead directly optimize for evading the detector with adaptive attacks~\cite{tramer2020adaptive}.
To minimize the chance of being detected without access to $\neg Det(x')$, the adversary thus makes an assumption: small perturbations are less likely to be detected than large ones.
The adversary then uses the perturbation size as a proxy for detectability and selects the smallest perturbation per observation among all successful perturbations.
This means that an operator has to assume the \emph{worst-case}, i.e., the smallest successful perturbation for each observation.

A difficulty for the adversary, and in turn also for the simulation, is to select a distance metric to measure perturbation size that correlates well with the chance of detection.
A common choice are $L_p$ metrics, but there is an active debate about which metrics align well with human perception, \citep[see for instance][]{Zhang2018-nq}.
With this in mind, we can now introduce a formal definition of the attack strategy:
\begin{definition}[Attack Strategy]
An attack strategy is a function 
\begin{equation*}
    \adv(x) \! = \!
    \begin{cases}
        \argmin\limits_{x' \in Cand_{\A}^{Succ}(x, M) } \mkern-26mu d(x', x) \text{,} & \! \! \! \text{if } Cand_{\A}^{Succ}(x, M) \! \neq \! \emptyset \\
        \quad\quad x & \! \! \! \text{otherwise},
    \end{cases}
\end{equation*}
of type $\adv \colon \mathcal{X} \rightarrow \mathcal{X}$ that returns the smallest perturbation obtained by applying a set of applicable adversarial attacks $\A$ to an observation $x$ given a distance metric $d$.
\end{definition}

We say an attack strategy is \emph{successful} if $\adv(x) \neq x$.
Since an attack strategy relies on empirical evaluations, it yields an upper bound on the minimum perturbation required for any attack strategy to be successful.

\begin{definition}[Smallest Upper Bound on Perturbation Size]
    The smallest upper bound on the perturbation size obtained by a successful attack strategy is
    \begin{equation*}
        \dsa(x, M) = 
        \begin{cases}
            d(\adv(x), x)\text{,} & \! \! \text{if }  \adv(x) \neq x \\
            \infty & \! \! \text{otherwise},
        \end{cases}
    \end{equation*}
    where $d$ is a distance metric on $\mathcal{X}$.
\end{definition}

{
\setlength{\algomargin}{1.5em}
\SetAlgoNlRelativeSize{0}
\begin{algorithm2e}[t!]
	\setstretch{1.13}
	\small
	\DontPrintSemicolon
	\InOut
	\Input{$\mathbf{X}$, $M$, $\A$, $Det$, $d$}
	\Output{$\pdamhat$}
    $r \gets 0$\;
    \BlankLine
        \For(\Comment*[f]{Outer Loop}){$x \in \mathbf{X}$}{
        \label{alg:dmc:outer-loop}
        $d_{min} \gets \infty$\;
        \For(\Comment*[f]{Inner Loop}){$a \in \A$}{ \label{alg:dmc:inner-loop}
            $x' \gets a(x, M)$\;
            \If{$M(x') \neq M(x) \land d(x', x) < d_{min}$}{ \label{alg:dmc:eval-success}
                $x'' \gets x'$, $d_{min} \gets d(x', x)$\;
            }
            }
        \If{$x'' \neq x$ and $\neg Det(x'')$}{ \label{alg:dmc:eval-detection}
            $r \gets r + 1$\;
        }
    }
	\Return $\frac{r}{\cardinality{\mathbf{X}}}$\;
	\caption{\acrlong*{mcs} of $\pdam$}
    \label{alg:direct_monte_carlo}
\end{algorithm2e}
}

We can now define an algorithm to find an estimate for $\pdam$ by \acrlong*{mcs}.
\autoref{alg:direct_monte_carlo} illustrates the idea: iterate over observations in a sample $\mathbf{X}$ (\autoref{alg:dmc:outer-loop}) and individual attacks of an attack strategy $\mathcal{A}$ (\autoref{alg:dmc:inner-loop}), measure the attack success (\autoref{alg:dmc:eval-success}) and return the ratio of observations for which an attack strategy was successful, i.e., the ones that are not detected (\autoref{alg:dmc:eval-detection}).  
The result is an unbiased and consistent estimate of the probability of attack success for an attack strategy.

\subsection{Estimating the Probability of Detection}
\label{subsec:inferred_damage_prob}

\autoref{alg:direct_monte_carlo} reveals the two limitations that we introduced earlier.
First, samples sizes $\cardinality{\mathbf{X}}$ must be small if evaluating $Det(x')$ is costly (\emph{\limitA}).
This, in turn, means that the sample size and the number of attacks to evaluate the success of an attack strategy must be small.
The small sample sizes limit the quality of the estimate.
Second, the evaluation of $Det$ occurs \emph{after} generating $x'$.
Since $x'$ depends on $M$, the result of the estimation is model-dependent (\emph{\limitB}).
One must repeat the estimation of $\pdam$ for each model.
We now show how one can overcome both limitations.

\subsubsection{Overcoming \limitA} 
One way to overcome \emph{\limitA} is to substitute the detector with a surrogate function that is inexpensive to evaluate.
Specifically, if there is a function $F$ that substitutes $\neg Det$, such that $F(x) = \neg Det(\adv(x))$, one can use $F$ instead of $\neg Det$ to evaluate the detector during a \acrlong*{mcs}.
We can express this in commutative diagram notation as
\begin{equation*}
\begin{tikzcd}
    \adv(x) \arrow[rr, "\neg Det"'] \arrow[rr, "F", dashed, bend left] &  & \neg Det(\adv(x))
\end{tikzcd}  
\end{equation*}
Finding a suitable $F$ is difficult since the domain of $F$ is the high-dimensional observation space.
In particular, estimating a function in a high-dimensional space may still require large sample sizes.
However, under the assumption that the detection probability correlates well with the perturbation size, one can first map $x$ to a distance and then use the distance as the domain of the detection function.
With this assumption, the commutative diagram changes to 
\begin{equation*}
\label{diag:f'}
\begin{tikzcd}
    & \dsa(x) \arrow[rrd, "F'", dashed] &  &                    \\
x \arrow[r] \arrow[ru] & \adv(x) \arrow[rr, "\neg Det"'] &  & \neg Det(\adv(x))
\end{tikzcd}
\end{equation*}
where $F'$ is a function of type $F'\colon \mathbb{R} \rightarrow \{0,1\}$, and the image of $d_\mathcal{A}$ the distance space induced by a metric $d$.

We further say that such an $F'$ has the \emph{point-wise detection commutation property} if
\begin{equation*}
    \forall x \in \mathcal{X}\colon Succ(\adv(x)) \Rightarrow F'(\dsa(x)) = \neg Det(\adv(x)) \eqdot
\end{equation*}
Such an $F'$ may not exist.
Think of two observations $x, \tilde{x} \in \mathcal{X}, x \neq \tilde{x}$ with $x'=\adv(x)$ and $\tilde{x}'=\adv(\tilde{x})$, where $d_\mathcal{A}(x) = d_\mathcal{A}(\tilde{x})$ but $\neg Det(x') \neq \neg Det(\tilde{x}')$, i.e., both observations have the same upper bound on the perturbation size.
Then, there is no $F'$ that has the detection commutation property.

Fortunately, we do not require a point-wise detection commutation.
Since $\pdamhat$ is a probabilistic estimate, it suffices that $F'$ is equivalent to the probability of detection.
\begin{definition}[Probabilistic Detection Commutation Property]
    \label{def:probabilistic-detection-commutation-property}
    We say a function $F': \R \rightarrow \R$ has the
    \emph{probabilistic detection commutation property}
    if
    \begin{equation}
        \label{eq:prob_det_comm_prop}
    P_{x \sim \mathcal{X}'}(\neg Det(\adv(x))) = \E_{x \sim \mathcal{X}'}(F'(\dsa(x))),
    \end{equation}
    where $ \mathcal{X}' = \{x \in \mathcal{X}: Succ(\adv(x))\}$.
\end{definition}

\noindent
Given an $F'$ that has the probabilistic detection commutation property, one can obtain $\pdam$ using the cumulative distribution function of $\dsa$, the \emph{Attack Success Distribution}.
\begin{definition}[Attack Success Distribution]
\begin{equation}
   \text{ASD}_{\mathcal{A}, M}(\tau) \coloneqq P_{x \sim Adv^M(\mathcal{X})}(\dsa(x, M) \! \leq \tau), \tau \geq 0
\end{equation}
\end{definition} 
Formally, this leads to the following theorem.
\begin{theorem}\label{theorem:prob_det_comm_to_pdam_calc}
    Let $F': \R \rightarrow \R$ be a function that fulfills the probabilistic detection commutation property.
    Then
    \begin{equation*}
        \pdam = \int_0^{\infty}F'(\tau) \ d\text{ASD}_{\mathcal{A}, M}(\tau) \ d\tau .
    \end{equation*}
\end{theorem}
\emph{Proof.} See \autoref{appendix:proof-theorem:prob_det_comm_to_pdam_calc}.

Using \autoref{theorem:prob_det_comm_to_pdam_calc}, the key to an estimation of $\pdam$ is an estimation of $\text{ASD}_{\mathcal{A}, M}$. Given a sample $\mathbf{X} \sim \mathcal{X}$, one can approach the Attack Success Distribution with the following empirical distribution function \citep[cf.][]{Dong2020-tr}.
\begin{definition}[Attack Success Ratio]
    \begin{equation*}
        \text{ASR}_{\mathcal{A}, M}(\tau) = 
        \frac{\cardinality{ \{x \in \mathbf{X} \mid \dsa(x, M) \leq \tau \} }}{\cardinality{\mathbf{X}}}
    \end{equation*}
\end{definition}
Empirical distribution functions are unbiased and consistent estimators of their distribution functions.
Applying this general property to $\text{ASR}_{\mathcal{A}, M}$ allows us to formulate an unbiased and consistent estimator for $\pdam$.
\begin{theorem}\label{theorem:prob_det_comm_to_pdam_est}
    The estimator
    \begin{align*}
        \pdamhat &= \int_0^{\infty}F'(\tau) \ d\text{ASR}_{\mathcal{A}, M}(\tau) \ d\tau \\
        &= \frac{1}{\cardinality{\mathcal{X}}}\sum_{x \in \mathcal{X}, \dsa(x) \neq \infty} F'(\dsa(x)) \label{eq_line:sample_mean}\tag{\Cross} 
    \end{align*}
    is an unbiased, consistent estimator of $\pdam$.
\end{theorem}
\emph{Proof.} See \autoref{appendix:proof-theorem:prob_det_comm_to_pdam_calc}.

\subsubsection{Overcoming \limitB}

What is left to discuss is how to construct an $F'$ that fulfills the probabilistic commutation property.
A natural choice is to use the probability of detection conditioned on~$\tau$.
Intuitively, this probability is the ratio of observations with $d_{\mathcal{A}}(x)=\tau$ that one expects to be detected.
Formally, we define this as a probability function.
\begin{definition}[Detection Probability Function]
\begin{equation*}
        \Dma(\tau) = P_{x \sim \mathcal{X}}(\neg Det(\adv(x)) \mid \ \dsa(x) = \tau)
\end{equation*}
\end{definition}
With a suitable distance measurement, we can assume that the perturbation size statistically determines the detection probability function.
In this case, the detection probability \emph{does not depend on the choice of a model}, i.e.,
\begin{equation*}
    \begin{split}
        \Dma[M_1](\tau) &= P_{x \sim \mathcal{X}}(\neg Det(\adv[M_1](x)) \mid \ \dsa(x) = \tau) \\
        &= P_{x \sim \mathcal{X}}(\neg Det(\adv[M_2](x)) \mid \ \dsa(x) = \tau) \\
        &= \Dma[M_2](\tau)\eqdot
    \end{split}
\end{equation*}
where $M_1 \neq M_2$.
One can show that $\Dma$ is indeed a suitable choice for $F'$.
\begin{theorem}\label{theorem:prob_det_func_has_prob_det_comm}
    The \emph{Detection Probability Function} has the probabilistic detection commutation property.
\end{theorem}

\emph{Proof.} See \autoref{appendix:proof-theorem:prob_det_func_has_prob_det_comm}.

\noindent
Combining \autoref{theorem:prob_det_func_has_prob_det_comm} with
\autoref{theorem:prob_det_comm_to_pdam_calc} gives
\begin{corollary}
	\label{cor:pdam_by_dma_csr}
    The probability of damage $\pdam$ is
    \begin{equation*}
        \pdam = \int_0^{\infty}\Dma(\tau) \ d\text{ASD}_{\mathcal{A}, M}(\tau) \ d\tau
    \end{equation*}
    Furthermore, the estimator
    \begin{align*}
        \pdamhat &= \int_0^{\infty}\Dma(\tau) \ d\text{ASR}_{\mathcal{A}, M}(\tau) \ d\tau \\
        &= \frac{1}{\cardinality{\mathcal{X}}}\sum_{x \in \mathcal{X}, \dsa(x) \neq \infty} \Dma(\dsa(x)) 
    \end{align*}
    is an unbiased, consistent estimator of $\pdam$.
\end{corollary}

\noindent
A useful implication of Corollary~\ref{cor:pdam_by_dma_csr} is that one can further refine $\Dma$ to include prior knowledge and assumptions in the calculation of $\pdam$.
For instance, recall that an adversary uses an attack strategy $\adv$ to find the smallest perturbation for an observation.
The rationale is that the adversary expects that the chances of being detected increase with the perturbation size.
With this assumption, one can simplify $\Dma$ to monotonic non-decreasing functions or even a logistic curve.
One must then only query the detector with a data sample and then fit a posterior for $\Dmahat$, e.g., with Bayesian estimation, see \autoref{appendix:sketch-dma}.
A further takeaway from this section is that \emph{the estimate $\Dmahat$ depends on the application but not on a specific model}.
Think of our upload filter example.
Upload filters are used in different applications, e.g., to detect copyright infringement of portrait photographs (Application~A) and violent content in pictures (Application~B).
For each application, one must estimate a detection probability function: $\Dmahat^{photo}$ for Application~A and $\Dmahat^{violent}$ for Application~B.
However, within Application~A, $\Dmahat^{photo}$ can be used to estimate the risk for different copyright detection models, e.g., trained with different hyperparameter settings.
Likewise, $\Dmahat^{violent}$ can be used to estimate the risk of different violence detection models.
This reduces the number of detection probability functions one has to fit by querying a detector from one per model to one per application.
Thus, the independent estimation of the detection probability function mitigates both \emph{\limitA} and \emph{\limitB}, and hence reduces the effort for risk estimation.

\subsection{Estimations without a Detector}
\label{sec:relative-estimation}

In some cases, collecting a sample from a detector to fit $\Dma$ is infeasible.
For instance, think of academic benchmarks that compare adversarial attacks or defense methods.
There, querying a human detector is often beyond the scope of the study.
One may not even have an indication of which magnitude of perturbation size is actually required for a detection to be successful.
In such a case, selecting and estimating a suitable detection probability function is not possible.

However, one can still obtain a \emph{relative comparison of the operational risk} between models.
One way is to assume that the average sensitivity of models to adversarial perturbations is similar to the sensitivity of a potential detector.
Formally, this gives an average detection function
\begin{equation}
\label{eq:damhatavg}
    \Dmahat[M]^{avg}(\tau)= 1 - \frac{1}{J} \sum_{j=1}^{J} ASR_{\A, M_j}(\tau)
\end{equation}
where $M_1, M_2, \dots, M_j, \dots, M_J$ are the models to compare.
Intuitively, using $\Dmahat[M]^{avg}$ as a detection function means that models more sensitive to adversarial examples than the average will obtain a high operational risk.

If all $ASR_{\A, M_J}$ estimates are based on the same sample $\mathbf{X}$, one can rearrange \autoref{eq:damhatavg} to make its computation efficient.
We first define $W(\tau)$ to count the combination of observations and models where the adversarial example with the smallest perturbation size is further away than $\tau$.
\begin{equation*}
    W(\tau) = \cardinality{\{(i, j) \mid i \in [I], j \in [J], \dsa(x_i, M_j) > \tau \}} \eqdot
\end{equation*}

We then have

\resizebox{\linewidth}{!}{
\begin{math}
	\begin{aligned}
	        \Dmahat^{avg}(\tau) &= 1 - \frac{\sum_{j = 1 \dots J}ASR_{\A, M_j}(\tau)}{J} \\
	        &= \frac{1}{J} \left(\sum_{j = 1 \dots J} (1 - ASR_{\A, M_j}(\tau))\right) \\
	        &= \frac{1}{\cardinality{\mathbf{X}} \cdot J} \left(\sum_{j = 1 \dots J} (\cardinality{\mathbf{X}} - \cardinality{\{x \in \mathbf{X} \mid \dsa(x)\leq \tau\}})\right) \\ 
	        &= \frac{1}{\cardinality{\mathbf{X}} \cdot J} \sum_{j = 1 \dots J} \cardinality{\{x \in \mathbf{X} \mid \dsa(x) > \tau\}} \\
	        &= \frac{1}{\cardinality{\mathbf{X}} \cdot J} W(\tau)
	\end{aligned}
\end{math}%
\vspace{0.3em}
}

With Corollary~\ref{cor:pdam_by_dma_csr}, we have 
\begin{equation}
    \label{eq:relative-pdam}
    \pdamhat = \frac{1}{\cardinality{\mathbf{X}}^2 \cdot J} \sum_{j = 1 \dots J} W(\dsa(x_i, M_j))
\end{equation}
\autoref{alg:zero_knowledge_pdam} in \autoref{appendix:algorithm-pdam-without-detector} summarizes the estimation of $\pdamhat$ using \autoref{eq:relative-pdam}.
The algorithm helps identify the model with the lowest probability of damage even if collecting a sample from a detector is infeasible.

\section{Experiments}

\begin{figure}[t!]
	\centering
	\includegraphics[width=0.9\columnwidth]{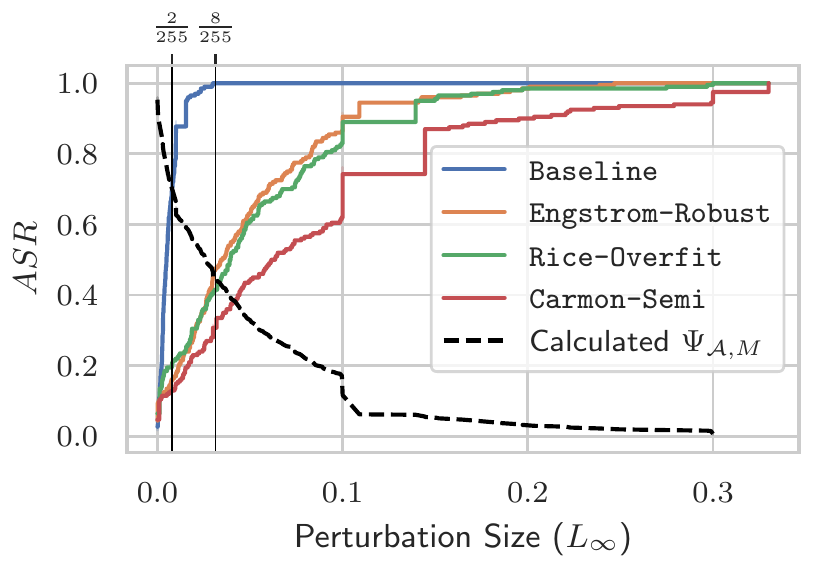}
	\vskip -0.1in
	\caption{
		\emph{Solid, colored:} ASR of the different models on 200 observations.
		\emph{Dashed, black:} $\Dma$ estimated according to \autoref{sec:relative-estimation}.
		\emph{Solid, black:} Vertical lines marking $L_\infty \in \{\frac{2}{255}, \frac{8}{255}\}$.
	}
	\label{fig:experiments}
\end{figure}

This section demonstrates that $\pdamhat$ provides a consistent and unbiased evaluation to compare the robustness of machine learning models without the need to choose a threshold on the perturbation size.
Our experimental setup is representative of how academic benchmarks comparing adversarial robustness are typically constructed.
Our goal is to underline the usefulness of our metric in common setups using open-source models and attack implementations.
Hence, our choice of models, attacks, and parametrization is arbitrary and can be replaced with any other use-case.\footnote{Our implementations and results are available at \url{https://github.com/duesenfranz/risk_scores_paper_code}.}

\begin{figure*}[t!]
\begin{center}
            {\includegraphics[width=0.7\textwidth]{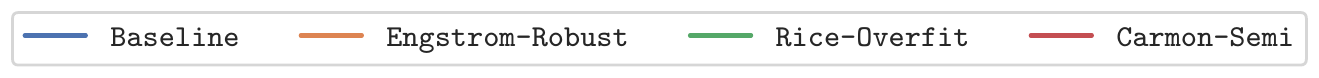}
            \vspace{-0.5em}}
        \subfigure[MPS with increasing sample size.]{\label{fig:stability:vulnerability}\includegraphics[width=0.46\textwidth]{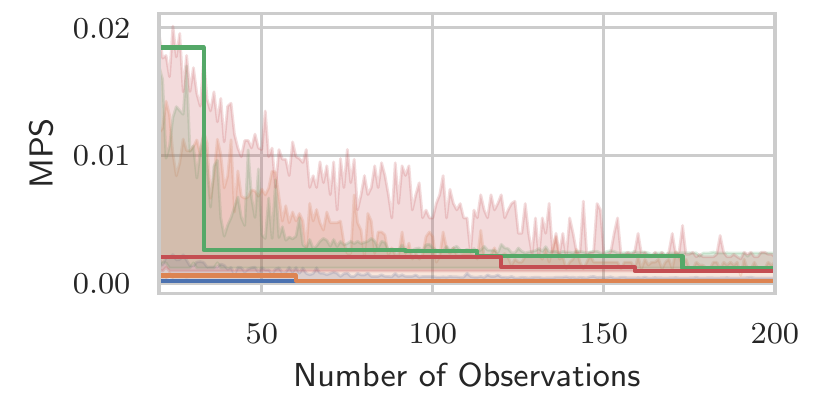}}
	\hfill
	\subfigure[$\pdamhat$ with increasing sample size.]{\label{fig:stability:pdam}\includegraphics[width=0.46\textwidth]{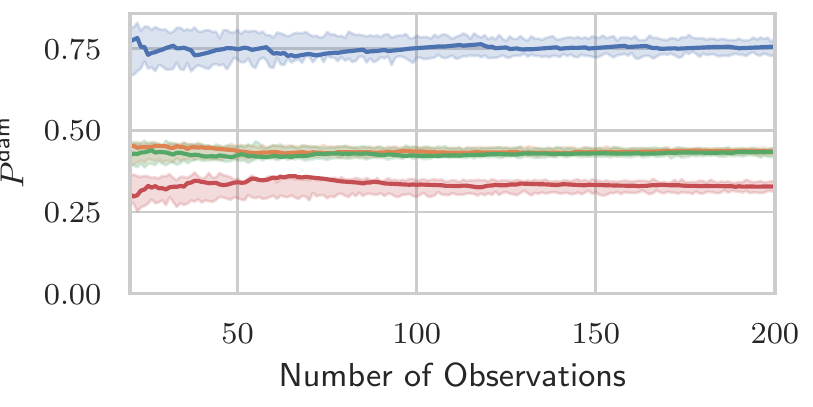}}
	 \caption{Convergence of robustness statistics.
	 	Each line plots a statistic measured on $n \in [20, 200]$ observations.
	 	The error band is the $5\%$ and the $95\%$ percentile, calculated by sampling $50$ times with replacement.}
	 \label{fig:stability}
\end{center}
\vskip -0.2in 
\end{figure*}

\begin{table}[t!]
	\centering
 	\caption{
		Summary statistics of the model robustness over 200 observations.
		Smaller values are better for all metrics but MPS.
            The best values are highlighted in bold.
	}
        \vskip 0.1in 
	\resizebox{\columnwidth}{!}{
		\begin{tabular}{lllll}
\toprule
                  Model &     $\widehat{{P}}^\text{dam}$ &          $\text{ASR}\left(\frac{2}{255}\right)$ &          $\text{ASR}\left(\frac{8}{255}\right)$ &                               MPS \\
\midrule
\texttt{\modelBaseline} &                           0.76 &                           0.70 &                           1.00 &                           0.00018 \\
\texttt{\modelEngstrom} &                           0.44 &                           0.16 &                           0.48 &                           0.00020 \\
    \texttt{\modelRice} &                           0.43 &                           0.20 &                           0.42 & \fontseries{b}\selectfont 0.00119 \\
  \texttt{\modelCarmon} & \fontseries{b}\selectfont 0.33 & \fontseries{b}\selectfont 0.13 & \fontseries{b}\selectfont 0.33 &                           0.00095 \\
\bottomrule
\end{tabular}

	}
	\label{fig:summary_stats}
        \vskip -0.1in 
\end{table}

\subsection{Setup}
We compare the operational risk of publicly available models on CIFAR-10~\cite{krizhevsky2009learning}.

\emph{Models.} The list of models includes a baseline classifier and three other models trained to achieve high adversarial robustness.
We obtained all models from public repositories.\footnote{
	All models can be downloaded using RobustBench~\cite{croce2021robustbench}.
}
\begin{enumerate}[itemsep=-0.5em, topsep=0em, align=left]
    \item[\modelBaseline] \cite{croce2021robustbench}: A baseline model trained without a specific focus on robustness.
    \item[\modelCarmon] \cite{carmon2019unlabeled}: A model trained for robustness with a semi-supervised learning method.
    \item[\modelEngstrom] \cite{engstrom2019adversarial}: A model trained for robustness by  adversarial training.
    \item[\modelRice] \cite{rice2020overfitting}: A model trained for  robustness by a combination of adversarial training and a focus on minimizing overfitting.
\end{enumerate}

\emph{Attacks.} We define an attack strategy based on a set of attacks $\mathcal{A}$ based on Foolbox~\cite{rauber2017foolboxnative}, an open-source attack library.
We use Projected Gradient Descent (PGD), PGD with Adam optimizer, and DeepFool.

\emph{Parametrization.} We use RobustBench~\cite{croce2021robustbench} to run attacks on observations in the test set.
We instantiate each attack with eight different values for epsilon, i.e., $\cardinality{\mathcal{A}}=24$, set $d = d_{\infty}$, and consider a perturbation successful if the model prediction does not agree with the ground truth.

\emph{Metrics.} Next to our metric $\pdamhat$, we compute three alternative metrics.
We compute \emph{$\text{ASR}\left(\tau\right)$}, with two thresholds $\tau \! \in \! \{\frac{2}{255},\! \frac{8}{255}\}$:
the fraction of observations with at least one successful adversarial example with perturbation size $L_\infty \leq \tau$.
$\text{ASR}\left(\tau\right)$ is the maximum likelihood estimator for the adversarial risk using the $0$-$1$ loss (cf.\ \autoref{sec:related-work}).
The threshold $\frac{8}{255}$ is arbitrary but common \citep[e.g., see][]{rice2020overfitting}.
Another common approach is to estimate the population parameters of the perturbation sizes $\dsa(x), x \in \X$, \citep[e.g., see][]{Carlini2019-yj}.
This approach is not reliable since estimates are either not robust to outliers (e.g., the average over $\dsa(x)$, see \autoref{sec:stability_average_perturbation_size}) or biased by nature (e.g., the median).
To demonstrate the issue of relying on perturbation size as a metric, we compute the size of the smallest perturbation that alters the model prediction, the \emph{Minimal Perturbation Size (MPS)}.
MPS is non-robust to outliers and biased.
\subsection{Results}
\label{sec:experiments-results}

\autoref{fig:experiments} plots the attack success rate (ASR) against the perturbation size.
Since this is an academic benchmark, we do not have access to a human detector in the loop.
Hence, we proceed as outlined in \autoref{sec:relative-estimation} to calculate $\Dma$.
Based on \autoref{fig:experiments}, an operator would choose \modelCarmon, the model with the lowest ASR for all perturbation sizes.
The second best is \modelRice since the ASR (green line) is lower than the one of \modelEngstrom (yellow line) on most of the perturbation spectrum.
The \modelBaseline performs poorly:
an adversary can find successful perturbations for each observation even for a small perturbation budget.

\autoref{fig:summary_stats} summarizes the \emph{ASR} plot with the metrics $\pdamhat$, $\emph{\text{ASR}}\left(\frac{2}{255}\right)$, $\emph{\text{ASR}}\left(\frac{8}{255}\right)$ and \emph{MPS}.
The selection based on $\pdamhat$ corresponds to the visual inspection: \modelCarmon is the model with the lowest value ($\pdamhat=0.33$), followed by \modelRice ($\pdamhat=0.43)$.
A benefit of $\pdamhat$ is that it remains actionable even if there are too many models for a visual inspection.

The $ASR(\tau)$ metrics, however, contradict each other.
$\emph{\text{ASR}}\left(\frac{8}{255}\right)$ suggests that \modelRice is more robust than \modelEngstrom; $\emph{\text{ASR}}\left(\frac{2}{255}\right)$ suggests the opposite.
Further, $\emph{\text{ASR}}\left(\tau\right) = 1.0$ for all models if the perturbation size is unconstrained, i.e., $\tau \rightarrow \infty$.
Thus selecting any threshold on \emph{ASR} is arbitrary, and results are volatile.

\emph{MPS} suggests that \modelRice is the most robust model.
However, \emph{MPS} is biased: with increasing sample size, its value approaches the true size of the smallest adversarial example, which is close to $0$ for all models, see \autoref{fig:stability:vulnerability}.
Indeed, if adversaries can choose from an infinite number of observations, they likely find an adversarial example with a very small perturbation.
\emph{MPS} has high variance for small sample sizes $n<100$, i.e., results are insignificant.
On the other hand, $\pdamhat$ is consistent and unbiased, i.e., the metric converges to its true expected value with increasing sample size, see \autoref{fig:stability:pdam}.

In summary, our experimental results confirm our theoretical analyses.
Neither $\emph{\text{ASR}}\left(\tau\right)$ nor \emph{MPS} are reliable metrics for selecting a robust model.
$\pdamhat$ allows for relative model robustness comparisons, even when no detector is available.

\section{Conclusions}

Estimating the damage caused by adversarial attacks is difficult.
Standard \acrlong*{mcs} is inefficient because it is limited to small sample sizes and model-specific estimates.
A consequence is that resulting metrics do not give reliable estimates of model robustness.
This prevents operators of machine learning models from translating high-level regulations on AI safety and security into actionable technical requirements.

In this article, we put forward an original approach to quantifying the risk of adversarial attacks that overcomes current limitations.
To this end, we first decompose the damage caused by adversarial attacks into the probability that an attacker is successful and the probability that an attack goes undetected.
We then propose an unbiased and consistent estimator for both quantities.
For cases where one does not have access to a detector, we provide an alternative method that allows comparing the risk between models.
The results are interpretable statistical estimates that provide an empirical basis for operators to select the model with the least risk of damage from adversarial attacks.

\section*{Acknowledgments}
This work was partially funded by the German BMWI project KI-LOK.

\bibliography{bibliography.bib}
\bibliographystyle{icml2023}

\clearpage
\newpage
\appendix
\section*{Appendix}
\label{app:appendix}

\section{Proof of \autoref{theorem:prob_det_comm_to_pdam_calc} and \autoref{theorem:prob_det_comm_to_pdam_est}}
\label{appendix:proof-theorem:prob_det_comm_to_pdam_calc}

\begin{proof}
    Starting with \autoref{eq:pdam_by_test_distribution}, we have
    \begin{align*}            
        \pdam &=  P_{x \sim \mathcal{D}}(Succ(\adv(x)),\ \neg Det(\adv(x))) \\
        & = P_{x \sim \mathcal{D}}( \neg Det(\adv(x)) \mid Succ(\adv(x))) \\
        & \quad \times P_{x \sim \mathcal{D}}(Succ(\adv(x))) \\
        &= \E_{x \sim \mathcal{D}}( F'(\dsa(x)) \mid Succ(\adv(x))) \\
        & \quad \times P_{x \sim \mathcal{D}}(Succ(\adv(x))) \\
        &= \E_{x \sim \mathcal{D}}( \begin{cases}
            F'(\dsa(x))\text{,} & \! \! \text{if } \dsa(x) < \infty \\
            0 & \! \! \text{otherwise}
        \end{cases}) \label{eq_line:pdam_as_expectancy}\tag{\textasteriskcentered} \\
        &= \int_0^{\infty}F'(\tau) \ d\text{ASD}_{\mathcal{A}, M}(\tau) \ d\tau,
    \end{align*}
    where the last line is because $\text{ASD}$ is the distribution function of $\dsa$ and because of the Law of the Unconscious Statistician (LOTUS).
    What remains to be proven is the unbiasedness and consistency of $\pdamhat$
    and the equality in Line~\ref{eq_line:sample_mean}.
    The latter is a direct result of the equality
    \begin{equation*}
        \text{ASR}_{\mathcal{A}, M} =  \frac{1}{\cardinality{\mathcal{X}}}\sum_{x \in \mathcal{X}, \dsa(x) \neq \infty} \chi_{[\dsa(x), \infty)},
    \end{equation*}
    where $\chi$ denotes the characteristic function.
    For unbiasedness and consistency, note that $\pdamhat$ in Line~\ref{eq_line:sample_mean} is the sample mean of the expected value in Line~\ref{eq_line:pdam_as_expectancy}, and therefore an unbiased and consistent estimator of $\pdam$.\qedhere
\end{proof}

\section{Proof of \autoref{theorem:prob_det_func_has_prob_det_comm}}
\label{appendix:proof-theorem:prob_det_func_has_prob_det_comm}

\begin{proof}
	We have to prove that
	\begin{equation*}
		P_{x \sim \mathcal{D}'}(\neg Det(\adv(x))) = \E_{x \sim \mathcal{D}'}(\Dma(\dsa(x)))\eqdot
	\end{equation*}
	With the Law of Total Probability, we have
	\begin{equation*}
		\begin{split}
			&P_{x \sim \mathcal{D}'}(\neg Det(\adv(x))) \\
			=\ &\E_{\tilde{x} \sim \mathcal{D}'}(\P_{x \sim \mathcal{D}'}(\neg Det(\adv(x)) \mid \ \dsa(x) = \dsa(\tilde{x}))) \\
			=\ &\E_{\tilde{x} \sim \mathcal{D}'}(\Dma(\dsa(\tilde{x}))) \\
			=\ &\E_{x \sim \mathcal{D}'}(\Dma(\dsa(x)))      \qedhere
		\end{split}
	\end{equation*}
\end{proof}

\section{Sketch of Fitting a Detection Probability Function $\Dma$ with Logistic Regression}
\label{appendix:sketch-dma}

\begin{figure}[t]
    \centering
    \includegraphics[width=\linewidth, height=12em]{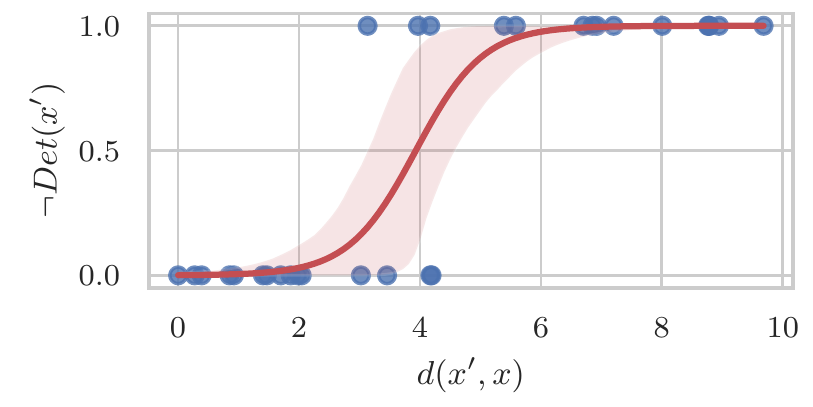}
    \caption{Sketch of fitting a detection probability function $\Dma$ with logistic regression. \
    \emph{Blue}: Scatterplot for $d(x', x) \!=\! d(a(x, M), x), a \in \mathcal{A}$ and $\neg Det(x')$. \
    \emph{Red}: Estimate $\Dmahat$ by logistic regression.}
    \label{fig:logistic_regression_dma}
\end{figure}

\autoref{fig:logistic_regression_dma} sketches how a detection probability function can be
estimated with logistic regression using only $30$ samples.
\section{Algorithm for $\pdamhat$ without a Detector}
\label{appendix:algorithm-pdam-without-detector}

{
\setlength{\algomargin}{1.5em}
\SetAlgoNlRelativeSize{0}
\begin{algorithm2e}[t]
	\setstretch{1.13}
	\small
	\DontPrintSemicolon
	\InOut
	\Input{$x_1, \dots, x_I$, $M_1, \dots, M_J$, $\A$, $d$}
	\Output{$\pdamhat$ for $M_1, M_2, \dots, M_J$}
	
	$W \gets []$\;
	$ASR \gets []$\;
	$\pdamhat \gets []$\;
	\BlankLine
	\For(\Comment*[f]{Part A}\label{alg:algorithm-pdam-without-detector:A}){$j = 1 \dots J$}{
		$d_{min} \gets \infty$\;
		$ASR[j] \gets []$\;
		\For{$i = 1 \dots I$}{
			\For{$a \in \A$}{
				$x' \gets a(x_i, M_j)$\;
				\If{$M(x') \neq M_j(x_i) \land d(x', x_i) < d_{min}$}{
					$x'' \gets x'$\;
					$d_{min} \gets d(x', x)$\;
				}
			}
			\eIf{$x'' \neq x_i$}{
				$\text{push}(W, d(x'', x_i))$\;
				$\text{push}(ASR[j], d(x'', x_i))$\;
			}{
				$\text{push}(W, \infty)$\; 
			}
		}
	}
	$\text{sort\_descending}(W)$\; 
	\For(\Comment*[f]{Part B}\label{alg:algorithm-pdam-without-detector:B}){$j = 1 \dots J$}{
		$\pdamhat[j] \gets 0$\;
		\For{$\tau \in ASR[j]$}{
			$\pdamhat[j] \gets \pdamhat[j] + \text{index}(W, \tau)$\; 
		}
		$\pdamhat[j] \gets \frac{\pdamhat[j]}{I^2 \cdot J}$\;
	}
	\Return $\pdamhat$\;
	\caption{Estimation of $\pdam$ without Detector}
	\label{alg:zero_knowledge_pdam}
\end{algorithm2e}
}
\autoref{alg:zero_knowledge_pdam} calculates $\pdam$
if no detector is available, see \autoref{eq:relative-pdam}.
The algorithm consists of two parts:
\begin{enumerate}[itemsep=-0.5em, topsep=0em, align=left]
	\item[Part A] calculates the smallest perturbation sizes $\dsa(x)$ of the adversarial examples for all models and observations $x$
		and collects them in two arrays:
		\begin{enumerate}[itemsep=-0.5em, topsep=0em, align=left]
			\item A sorted array \texttt{W} which contains the smallest successful perturbations of all models.
			\item A double array \texttt{ASR}, which contains the smallest successful perturbations organized into subarrays per model.
		\end{enumerate}
	\item[Part B] computes $\pdamhat$ of each model $M_i$ by comparing
		\texttt{ASR[i]} with \texttt{W}:
		It sums the indices in \texttt{W} of the smallest successful perturbations for the model $M_i$.
		It then divides the result by the square of the number of observations and by the number of models.
\end{enumerate} 
Finally, the algorithm returns $\pdamhat$.

\begin{figure}[t!]
	\centering
	\includegraphics[width=0.9\columnwidth]{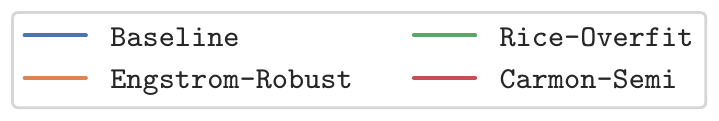}
	\includegraphics[width=0.9\columnwidth]{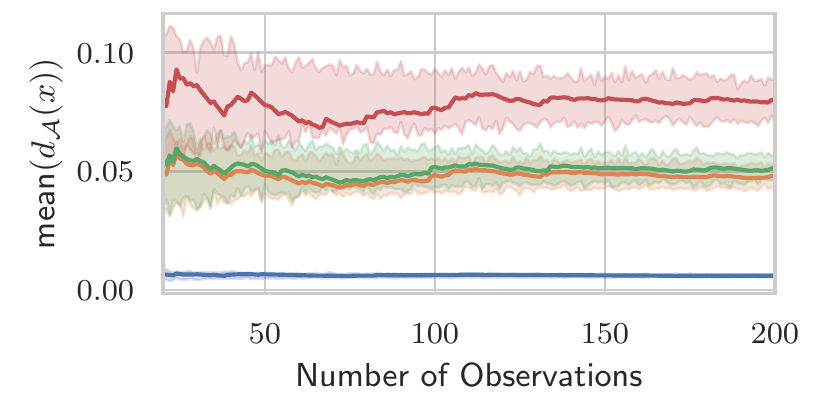}
	\caption{Convergence of the average perturbation size.
	Each line plots the average perturbation size, measured on $n \in [20, 200]$ observations.
	The error band is the $5\%$ and the $95\%$ percentile, calculated by sampling $n$ observations $50$ times with replacement.}
	\label{fig:average_perturbation_size}
\end{figure}

\section{Average Perturbation Size with Increasing Sample Size}
\label{sec:stability_average_perturbation_size}

\autoref{fig:average_perturbation_size} plots the average perturbation size calculated on a subset of the $200$ observations.
The non-robustness to outliers causes the wide error bands, which leads to an overlap of the percentiles of \modelCarmon and \modelRice for estimates based on fewer than $50$ observations.
In contrast, $\pdamhat$ yields statistically significant results even for estimates on fewer than $50$ observations, see \autoref{fig:stability}.

\end{document}